# INTRODUCTION TO RELATIONAL NETWORKS FOR CLASSIFICATION

Vukosi Marivate* and Tshilidzi Marwala**
University of the Witwatersrand
Johannesburg
South Africa
vukosi.marivate@ieee.org, t.marwala@ee.wits.ac.za

**ABSTRACT**

The use of computational intelligence techniques for classification has been used in numerous applications. This paper compares the use of a Multi Layer Perceptron Neural Network and a new Relational Network on classifying the HIV status of women at ante-natal clinics. The paper discusses the architecture of the relational network and its merits compared to a neural network and most other computational intelligence classifiers. Results gathered from the study indicate comparable classification accuracies as well as revealed relationships between data features in the classification data. Much higher classification accuracies are recommended for future research in the area of HIV classification as well as missing data estimation.

## 1. Introduction

Acquired immunodeficiency syndrome (AIDS) is a collection of symptoms and infections resulting from the specific damage to the immune system caused by the human immunodeficiency virus (HIV) in humans [1]. South Africa has seen an increase in HIV infection rates in recent years as well as having the highest number of people living with the virus. This results from the high prevalence rate as well as resulting deaths from AIDS [2]. Research into the field is thus strong and ongoing so as to try to identify ways into dealing with virus in certain areas. Thus demographic data are often used to class people living with HIV and how they are affected. Thus proper data collection needs to be done so as to understand where and how the virus is spreading. This will give more insight into ways in which education and awareness can be used to equip the South African population. By being able to identify factors that deem certain people or populations in higher risk, the government can then deploy strategies and plans within those areas so as to help the people.

This paper investigates the use of a new architecture for classification of HIV. This method is compared to the neural network method that has been used by Leke [3]. The work done in the field of HIV classification using demographics has mostly yielded results that have 50% - 70% accuracy. Improvement in the classification is needed to have a much better understanding of the underlying information in the data. An important factor though is the relation between the data features. Knowing how the features impact on each other will reveal factors that are highly dependent. The relations can then pave the way for better decision making by health care professionals in areas that are affected by HIV. This paper first gives a background of the methods to be compared in Section 2. Section 3 discuses the data collection for the HIV data. Section 4 discusses the methodology. Section 5 discusses the results and Section 6 is the conclusion.

## 2. Background

### 2.1 Neural Networks

Neural Networks are computational models that have the ability to learn and model systems. They have the ability to model non-linear systems [4]. The neural network architecture used in this paper is a multilayer perceptron (MLP) network [4] as shown in Figure 1. This has two layers of weights which connect the input layer to the output layer. The middle of the network is made up of a hidden layer. The general equation of the MLP neural network is shown below (1):

$$y_k = f_{outer}(\sum_{j=1}^{M} w_{kj}^{(2)} * f_{inner}(\sum_{i=1}^{d} w_{ji}^{(1)} x_j + w_{j0}^{(1)}) + w_{k0}^{(2)}) \quad (1)$$

Parameter $y_k$ is the output from the neural network, $x_j$ represents the inputs into the neural network and $w$ represents the different weights between the nodes in the neural network. For the classification network the $f_{outer}$ activation function is a logarithmic activation function and $f_{inner}$ is the hyperbolic tangent function. Training is done using the back-propagation algorithm [4]. The inputs into the neural network are demographic data from an antenatal survey while the output from the neural network is the HIV status of the subject at hand.

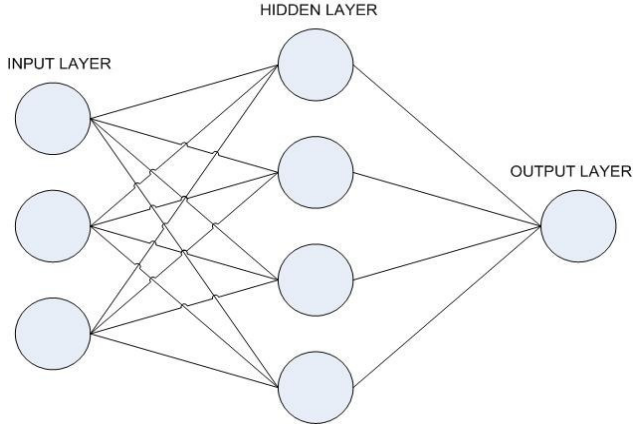

**Figure 1 MLP Neural Network**

## 2.3 Relational Network

The other network that is proposed in this paper and then used is the Relational Network. The network uses all the inputs as outputs from the system. This is what distinguishes the Relational Network from an extension neural network [5]. Each feature is taken as a node. All of the nodes are connected to each other with edges. Every node is connected to another node with two edges. The edges represent the relations of the one node on the other and visa versa. There is a weight assigned to an edge to quantify how connected the two corresponding nodes are. This is shown in an example in Figure 2.

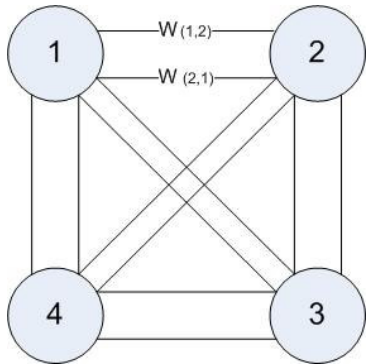

**Figure 2 Relational Network**

The general equation for all nodes in the network is:

$$x_k = \sum_{j=1}^{N} w_{kj} * f(x_j) \quad j \neq k \ for \ k \quad (2)$$

Here $w_{kj}$ represents the weights between the assigned to each function of another node $f(x_j)$. The functions used in this paper are:

- Linear - $f(x) = x$
- Logistic - $f(x) = \dfrac{1}{1+e^{-x}}$
- Hyperbolic Tangent - $f(x) = \dfrac{e^{2x}-1}{e^{2x}+1}$

Thus in this configuration all nodes are represented as weighted function of all other nodes. When training, the weights are adjusted so as to reduce the errors between true node values and calculated node values from (2). Thus the training of the network is supervised. The functions in the edges are used in order to find out how the data is related. By testing different network architecture one can gauge if the relations are linear or non-linear. The higher the weighting of the function of a node then the more dependent the two are. One node might be more dependent on the other; this will be noted by the weights. Thus this architecture can be viewed as a transparent architecture as the weights themselves can be analysed to explain the interconnections between the data being analysed. The training of the network and how it reduces its error is discussed in Section 4.2.

## 3. Data Collection and Pre-processing

The dataset that is used for this investigation is HIV data from antenatal clinics from around South Africa. It was collected by the department of health in the year 2001. The dataset contains multiple input fields that result from a survey. For example the provinces, region and race are strings. The age, gravidity, parity etc. are integers. Thus conversions are needed. The strings are converted to integers by using a look-up table e.g. there are only 9 provinces so 1 was substituted for a province etc.

Data collected from surveys and other data collection methods normally have outliers. These are normally removed from the dataset. In this investigation datasets that had outliers had only the outlier removed and the dataset was then classified as incomplete. This then means that the data can still be used in the final survey results if the missing values are imputed. The dataset with missing values was not used for the training of the computational methods. The data variables and their ranges are shown below in Table 1.

**Table 1. Features from Survey Data**

| Variable | Type | Range |
|---|---|---|
| Age | Integer | 1- 60 |
| Education | Integer | 0 - 13 |
| Parity | Integer | 0 - 15 |
| Gravidity | Integer | 0 - 11 |
| Age of Father | Integer | 1 - 90 |
| HIV Status | Binary | [0 , 1] |

Age is the age of the mother. Education is the level of education the mother has 1-12 being grades and 13 being tertiary education. Parity is the number of times the mother has given birth while Gravidity is the number of times the mother has been pregnant. Age of father is the age of the father and the last feature is the HIV status of the mother. The pre-processing of the data resulted in a reduction of training data. To use the dataset for training it needs to be normalised. This ensures that all the data variables can be used in training. If the data are not normalised, some of the data variables with larger variances will influence the result more than others. For an example, if we use Parity and Age data only the age data will be influential as it has large values. Thus all of the data are normalised between 0 and 1.

## 4. Methodology

The approach taken for this experiment is to build two classifiers and then compare their performances. Their performance is measured according to accuracy. The previous work carried out by Leke [3] is used as the base for the design of the MLP network. The Relational Network is then designed and the weights adjusted using the Metropolis Hastings algorithm.

### 4.1 Neural Network Training

The toolbox used for the neural network is the Netlab Toolbox [6]. The MLP was trained using the scaled conjugate gradient algorithm [6]. Education, Parity and Gravidity were used as binary numbers (4 binary digits each). Thus the number of inputs into the neural network is 14, this includes the normalised Age and Age of father. The output from the neural network is the HIV status. 1000 training cycles were used to train the neural network. It is important that the training data is unbiased. Most of the respondents of the survey are HIV negative thus if the neural network is trained with the data as is, it will most likely classify all data as negative as most of the data is negative. Thus the training data the number of HIV positive and negative respondents had to be the same. Thus in the training set the number of HIV positive sets can be re-used until we have an equal number of HIV positive.

### 4.2 Relational Network Training

To train the Relational Network a sampling algorithm was used. The sampling algorithm uses the mean square error as the basis of selection of the best solution. Initially weights are generated randomly. The error is calculated by comparing the real node input and the calculated value from the equation in the network. This is shown in equation 3:

$$e = \frac{1}{n}\sum_{k=1}^{n}(\bar{x} - x_k)^2 \qquad (3)$$

$\bar{x}$ is the training data, $x_k$ are the data outputs from the network as described in (2) and $n$ is the number of training sets. This is termed the mean square error. The algorithm works in the manner shown in Figure 3.

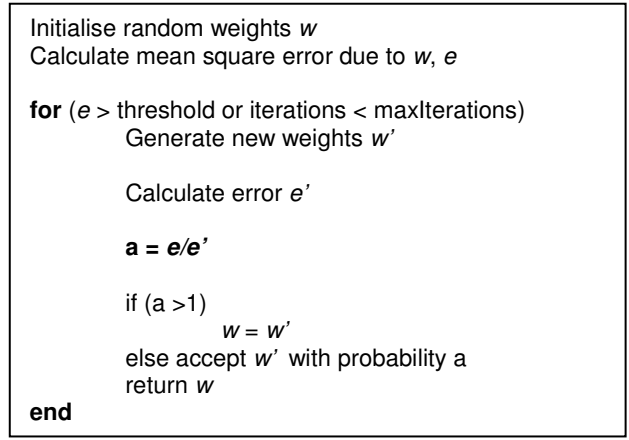

```
Initialise random weights w
Calculate mean square error due to w, e

for (e > threshold or iterations < maxIterations)
        Generate new weights w'

        Calculate error e'

        a = e/e'

        if (a >1)
                w = w'
        else accept w' with probability a
        return w
end
```

**Figure 3 Relational Network Training**

This is the Metropolis-Hastings algorithm [7] that is modified to sample for weights. The weights at the end of the algorithm have the least error. As can be noticed from this training method, the error used is the one that minimises the error of calculation for all features. This is different from an extension neural network [5]. The training chosen tries to build a general representation of all of the input features and not only for the data feature that needs to be classified. The training can be used to only focus on the feature that need be classified. All of the weights varied between 0 and 1.

## 5. Results

To test the classification methods a dataset that was not used in training was used. The testing dataset had 1500 sets from the survey that was unseen. The HIV classification was rounded as the output numbers from the networks ranged from 0 to 1. The outputs from the classifiers were then compared to the real outputs from the testing set and thus their performance could be measured.

### 5.1 Accuracy Measure

To measure the accuracy of the methods one cannot simply use the mean square error. The mean square error is a good measure to use for optimisation or sampling methods. In order to quantify the results, percentage accuracy is used as well as confusion matrices in order to be able to view how the classification is fairing especially for binary data such as HIV status. The confusion matrix [8] shows the cross-classification of the predicted class against the true class. Thus the confusion matrix indicates the number of true positives (*TP*), false positives (*FP*), true negatives (*TN*) and false negatives (*FN*). The total accuracy is measure as:

$$Accuracy(\%) = \frac{TP + TN}{TP + TN + FP + FN} \times 100 \qquad (4)$$

### 5.2 MLP Results

The best MLP configuration had 17 hidden nodes and the accuracy on the validation set was 55%. Thus the MLP could predict the HIV status 55% of the time.

**Table 2 MLP confusion matrix**

|  | Predicted positive | Predicted negative |
|---|---|---|
| True positive | 221 | 131 |
| True negative | 526 | 622 |

### 5.3 Relational Network Results

The Relational Network performance was trained with different activation functions in the edges. The best activation function was the linear activation which resulted in 60% classification accuracy. The logistic activation function resulted in an accuracy of 65.73%. The hyperbolic tangent activation resulted in 60% accuracy. The first confusion matrix is that of the linear function, shown in Table 3.

**Table 3. Linear function confusion matrix**

|  | Predicted positive | Predicted negative |
|---|---|---|
| True positive | 127 | 225 |
| True negative | 370 | 772 |

The total accuracy for the linear function is 60%. Its false positives are numerous but not as severe as the other functions. The second confusion matrix, for the logistic function, is shown in Table 4.

**Table 4. Logistic function confusion matrix**

|  | Predicted positive | Predicted negative |
|---|---|---|
| True positive | 77 | 275 |
| True negative | 239 | 909 |

The logistic function has a higher accuracy of 65.7%. This though is at a sacrifice of higher false positives. Thus the prediction that a person is HIV positive given their demographic is only correct 22% of the time. The final confusion matrix (Table 5) is that for the hyperbolic tangent function

**Table 5. Hyperbolic tangent function confusion matrix**

|  | Predicted positive | Predicted negative |
|---|---|---|
| True positive | 167 | 185 |
| True negative | 455 | 693 |

The hyperbolic tangent function has a total accuracy of 57%.

### 5.3 Weighting and Relations

The relational network allows for observation of the weights. By analysing the weights, the relations (linear or non linear) between the data features can be observed. The weights values varied between 0 and 1. For the linear function the largest weights connected to HIV were for age (0.3) while the least was 0.04 for Gravidity and 0.15 for Age of father. The logistic function had high weighting between HIV status and the age of the father. The hyperbolic tangent function had high weighting for the age (0.42) as well as the age of the father (0.25). Low weights were observed in all functions for gravidity and parity. The weights for these were in the order of 0.05. From observing the weights we can then setup questions like:

- The older the mother coming to the clinic the higher the chance of HIV?
- The older the father of the child the higher the HIV?

These questions then call for further investigation in the medical field.

## 6. Discussion

The results of the HIV classification vary with the different methods. The total accuracy of all the methods is compared in Figure 4.

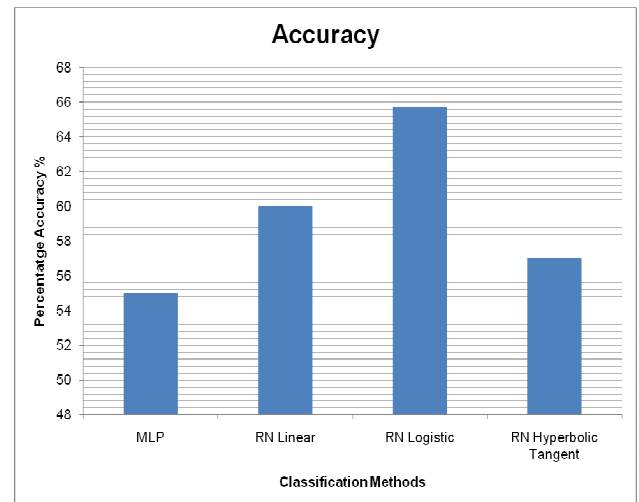

**Figure 4 Comparison of Accuracies**

The logistic relational network has the highest total accuracy. This though results in a high false negative count as compared to the other methods. When analysing the results of all of the methods a pattern emerges where if the total accuracy increases the false positives? This may be the result of the nature of the data itself. As discussed earlier to train the methods the data needs to be unbiased. The bias is due to the fact that close to 75% of the respondents of the survey is HIV negative. Thus if one predicts that a respondent is HIV negative they will be right 75% of the time. It seems that the higher the accuracy of the classification method the more the true negatives will increase while the true positives decrease. Thus a new goal recommended for researchers using this data is to have accuracies that are in the 90% range. The closer the accuracy is to 75% the higher the inaccuracies in predicting the HIV positive respondents.

Missing data imputation of HIV data has also been an area of research interest [9][10]. In this field the prediction of HIV has yielded results with total accuracies of up to 68% using methods such as Rough Sets [11], Auto-Associative Neural Networks [9] and Support Vector Regression [10]. These results are comparable to classification results. Thus further investigation on how the data itself impacts on the accuracy of the classification is crucial in improving the over-all results.

## 7. Conclusion

The use of a relational network to classify HIV has been discussed and the results are comparable to an MLP classification method. The results indicate that there are inter-relationships between the HIV data features. The HIV status is related to the age of the mother and the father. An increase in the classification accuracy of the method is an increase also in the number of false positive classification. This phenomenon occurs even though the training data is unbiased. Thus a challenge for future researchers is to get the classification accuracies for the HIV dataset to levels beyond 80% and close to the 90 percentile. The data itself is primarily composed (75%) of HIV negative respondents thus having accuracies higher than 80% are needed.